\documentclass[conference]{IEEEtran}
\IEEEoverridecommandlockouts
\usepackage{cite}
\usepackage{amsmath,amssymb,amsfonts,mathtools}
\usepackage{algorithmic}
\usepackage{graphicx}
\usepackage{textcomp}
\usepackage{xcolor}
\usepackage{tikz}
\usepackage{array}
\usepackage{comment}

\def\BibTeX{{\rm B\kern-.05em{\sc i\kern-.025em b}\kern-.08em
    T\kern-.1667em\lower.7ex\hbox{E}\kern-.125emX}}
 
 \newcommand*{\addheight}[2][.5ex]{%
   \raisebox{0pt}[\dimexpr\height+(#1)\relax]{#2}%
 }
 \newcommand{\choi}{Choi \textit{et al.}}
 \newcommand{\siamesethr}{Siamese-T-0.8}
 \newcommand{\siamesegau}{Siamese-G-0.95}

\newcolumntype{C}[1]{>{\centering\arraybackslash}m{#1}}

\DeclarePairedDelimiter\floor{\lfloor}{\rfloor}

\begin{document}

\title{Hue Modification Localization By Pair Matching}

\author{\IEEEauthorblockN{Quoc-Tin Phan}
\IEEEauthorblockA{\textit{DISI, University of Trento, Italy}}
\and
\IEEEauthorblockN{Michele Vascotto}
\IEEEauthorblockA{\textit{DISI, University of Trento, Italy}}
\and
\IEEEauthorblockN{Giulia Boato}
\IEEEauthorblockA{\textit{DISI, University of Trento, Italy}}
}

\maketitle

\begin{abstract}
Hue modification is the adjustment of hue property on color images. Conducting hue modification on an image is trivial, and it can be abused to falsify opinions of viewers. Since shapes, edges or textural information remains unchanged after hue modification, this type of manipulation is relatively hard to be detected and localized. Since small patches inherit the same Color Filter Array (CFA) configuration and demosaicing, any distortion made by local hue modification can be detected by patch matching within the same image. In this paper, we propose to localize hue modification by means of a Siamese neural network specifically designed for matching two inputs. By crafting the network outputs, we are able to form a heatmap which potentially highlights malicious regions. Our proposed method deals well not only with uncompressed images but also with the presence of JPEG compression, an operation usually hindering the exploitation of CFA and demosaicing artifacts. Experimental evidences corroborate the effectiveness of the proposed method.
\end{abstract}


\section{Introduction}\label{sec:intro}
Modern photography is losing its innocency due to the diversed use of image manipulation software, which allows even unexperienced users to modify digital images in different ways. Image contents are characterized mainly by geometric information like texture, edges and shapes, and by color information. Color modifications, even if does not effect geometric details, deceive human perception. They are very easy to be performed, and hard to be detected if implemented carefully.

\begin{figure}[h]
	\centering
	\begin{tabular}{ c c c c }
		\addheight{\includegraphics[width=17mm]{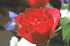}} & 
		\addheight{\includegraphics[width=17mm]{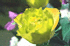}} &
		\addheight{\includegraphics[width=17mm]{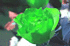}} & 
		\addheight{\includegraphics[width=17mm]{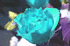}} \\
		$0^\circ$ &  $60^\circ$ & $120^\circ$ & $180^\circ$\\
	\end{tabular}
	\caption{Hue modification by different angles. $0^\circ$ means no modification. Better viewed in color.}
	\label{fig:hue_mod}
\end{figure}

 In this paper, we address the problem of local hue modification, which is defined as the adjustment of angular position on the color circle (or color wheel) within an image area. Figure \ref{fig:hue_mod} illustrates hue modification by different angles \footnote{https://www.imagemagick.org/Usage/color\_mods/, last access: 15/02/2019}.

 To cope with local image manipulations, previous works seek for artifacts of Color Filter Array (CFA) and camera sensor pattern noise. Vast camera sensors employ a CFA, where each sensor element captures the light at a  certain wavelength corresponding to a color component. The remaining color components at blind positions are interpolated from surrounding pixels. This interpolation is referred to as \textit{demosaicing}. Image manipulations will likely generate some local or global disturbances which are inconsistent to our ordinary demosaicing artifacts \cite{Ferrara2012}. More blind way to detect local disturbances is the extraction of statistical features of rich models capturing different types of neighboring dependencies \cite{Fridrich2012}. These features have been proved to be effective in manipulation detection and localization, see for instance \cite{Cozzolino2015,Li2017ForgeryLocalization}. Besides demosaicing, the imperfections of camera sensors also create sort of camera fingerprint, the so-called Photo-Response Nonuniformity (PRNU) noise, which is supposed to be present in every image \cite{Chen2008}. In the presence of manipulation, this pattern noise is distorted and this distortion can be exploited as a useful clue, provided that the reference PRNU can be reliably estimated and the forged region is sufficiently large.
 
 The specific local image manipulation considered in this paper, hue modification, distorts artifacts of demosaicing and neighboring dependencies. Based on this fact, the pioneering work in \cite{Choi2013} analyzes demosaicing artifacts and then estimates hue modification. Based on the observation that an interpolated value is bigger than the minimum and smaller than the maximum of its neighborhood, on the green channel the number of pixel values unsatisfying this condition should be the majority of pixel which are originally captured in this channel and a minority of interpolated pixels, resulting a big ratio between two quantities. The estimation of hue modification is done via searching over a set of modification angles until the aforementioned ratio is maximized. We want to point out that CFA analysis requires the knowledge of CFA configuration, at least the positions of green component. Such information is not always available, especially for online images. Moreover, when the image undergoes JPEG compression, demosaicing artifacts are significantly distorted. Differently, the method proposed in \cite{Hou2014} and \cite{Hou2017} recovers the modification angle, by modifying the questioned image with a set of angles and matching its residual with the reference PRNU. In real scenarios this technique is very difficult to be exploited since the assumption to know the reference PRNU (or have access to images to estimate it) is very strong and cannot be easily satisfied.
 
 In this work, we propose a novel method for detecting hue modification. Our methodology exploits the fact that two patches on the same image have the same inherent CFA configuration and demosaicing. Hue modification on a local region creates inconsistencies with the rest of the image, and thus pair-wise patch matching can reveal the forged region. To achieve such purpose, we propose a solution based on Siamese neural networks \cite{Bromley1993}, trained on positive pairs (two pristine patches) and negative pairs (a pristine and a modified patch). JPEG compression before and after hue modification is included during training, granting the network the capability to deal with real-world conditions. Finally, we fuse multiple outputs of patch matching to obtain a unique decision map (heatmap), on which a postprocessing is applied to precisely localize the forged region (Section II).
 Experiments demonstrate the effectiveness of the proposed approach (Section III).
\section{Proposed Method}\label{sec:method}
Hue modification is performed in the HSV space by adding an angle $\alpha$ to the value of $H$. Since defined on a circle, hue modification is periodic with a period $360^\circ$, i.e. a modification of $\alpha$ is identical to $\alpha \pm 360^\circ$. Besides hue, other attributes of a color in HSV space are saturation and value (brightness), whose changes are different from hue modification. Here, we investigate the detection of hue modification on: i) uncompressed images, ii) JPEG images where the modification is carried out before and after compression.

Given two rectangular patches $\mathcal{P}_i, \mathcal{P}_j$ of size $h \times w$ from the same image, we desire to estimate the logistic prediction $p_{ij}$ that two patches are inconsistent with respect to two corresponding modification angles $\alpha_i, \alpha_j$. The two patches are \textit{consistent} if $\alpha_i = \alpha_j = 0$ and \textit{inconsistent} if $\alpha_i \neq \alpha_j = 0$.

\begin{figure}[t]
	\includegraphics[width=\linewidth]{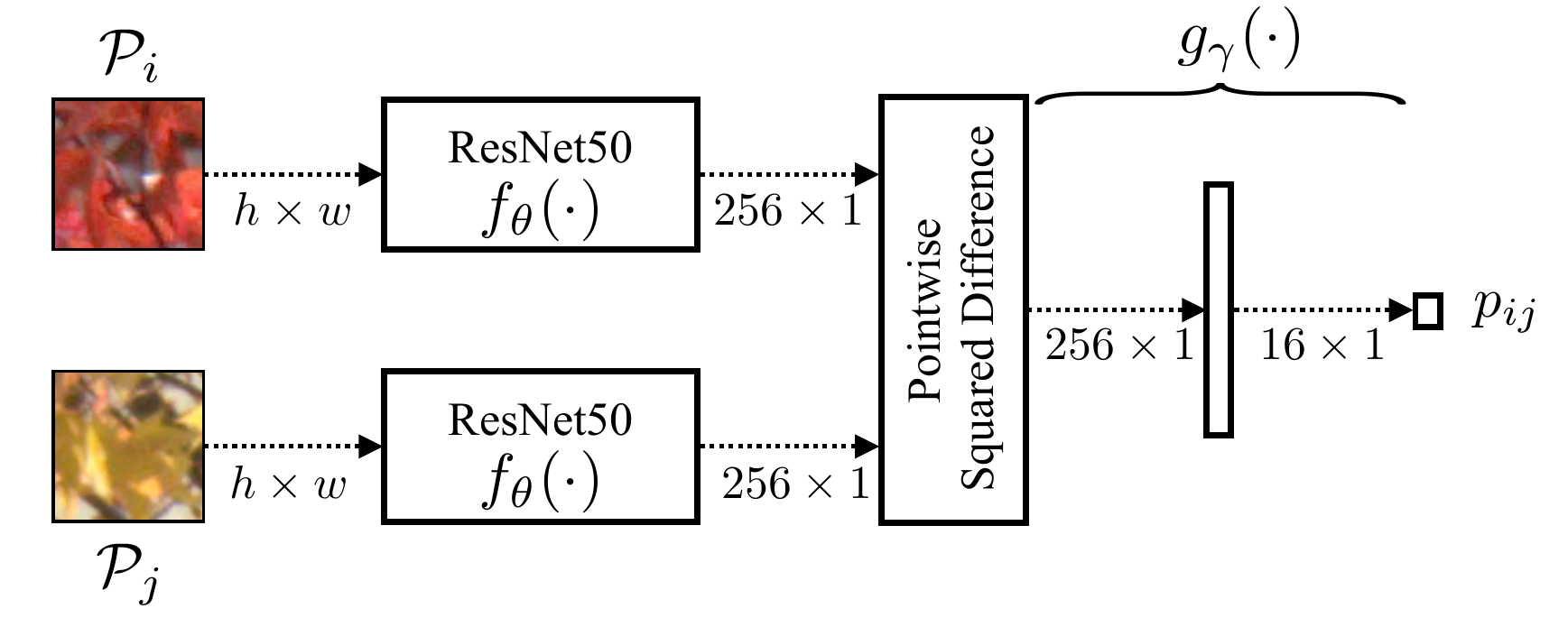}
		\vspace{-0.8cm}
	\caption{Proposed Siamese network architecture.}
	\label{fig:net_arch}
	\vspace{-0.2cm}
\end{figure}

We propose to verify the inconsistency of $\mathcal{P}_i$ and $\mathcal{P}_j$ by means of a Siamese neural network \cite{Bromley1993}. Siamese neural networks have been recently exploited for applications in multimedia forensics \cite{Cozzolino2018NoisePrint, Mayer2018, Huh2018}. This network architecture consists of two identical sub-networks $f_\theta$, followed by a non-linear classifier $g_\gamma$ that outputs an inconsistency score $z_{ij}$ whose standard logistic activation is defined as:
\begin{eqnarray}
p_{ij} = \frac{1}{1+e^{-z_{ij}}} \text{, and }
z_{ij} = g_\gamma \left ( \left [ f_\theta \left (\mathcal{P}_i \right ) - f_\theta \left (\mathcal{P}_j \right ) \right ]^2_{\text{pointwise}} \right) \nonumber \text{.}
\end{eqnarray}
The network parameters $\theta, \gamma$ are jointly optimized to minimize the binary cross-entropy of network logistic predictions $p_{ij}$ and patch inconsistencies $y_{ij}$, written in terms of a loss function over $N$ training patches:
\begin{equation}
	\mathcal{L} = \frac{1}{N^2-N}\sum_{i=1}^{N} \sum_{j=1,j \neq i}^{N} - y_{ij} \log(p_{ij}) - (1-y_{ij})\log(1-p_{ij}). \nonumber
\end{equation}
In Figure \ref{fig:net_arch}, we provide the sketch of the network architecture. We use the 50-layer Residual Network (ResNet50) \cite{He2016} as the feature extractor $f_\theta$, which outputs a $256$-dimensional feature vector. The inconsistency of features extracted from two patches are evaluated by a pointwise squared difference operator. The classifier $g_\gamma$ is a multilayer perceptron network composed by one hidden layer of $16$ units and one single-unit output layer with sigmoid activation outputting $p_{ij}$.
We train two separate Siamese networks end-to-end on large-scale synthetic training sets. The first model is trained on $400,000$ $64 \times 64$ patches extracted from uncompressed images from RAISE \cite{Dang-Nguyen2015} and Dresden \cite{Gloe2010}. To train the second model, we use the same $400,000$ patches and perform hue modification before or after JPEG compression with random quality factors in $[55,100]$. Parameters $\theta$ are initialized using ResNet50 pretrained on ImageNet \cite{Deng2009}. On each training iteration, we optimize the loss function with respect to $(\theta, \gamma)$ on a mini-batch of $64$ pairs, half of which is labeled as positive, i.e. both two patches are unmodified, and another haft is labeled as negative, i.e. one patch is modified by an angle randomly selected in $[30,330]$ with step 30, and its counterpart is unmodified. Hue modification and JPEG compression are carried out during training. We use Adam optimizer with the starting learning rate $10^{-4}$, and schedule to halve it every $5$ epochs after the first $30$ epochs until convergence.

\subsection{Detection and Localization}
\subsubsection{Heatmap creation}

The described architecture outputs the logistic patches inconsistency. Given a test image, we collect all inconsistency scores and generate a unique localization heatmap which potentially indicates malicious regions.

Let $H,W$ be height and width of the image, and $h,w$ be height and width of the small patch. By using a sliding window with stride $s$, the total number of patches will be $N = N_H \times N_W$, where $N_H = \floor*{\frac{H-h}{s}} + 1$ and $N_W= \floor*{\frac{W-w}{s}} + 1$ are number of patches along each dimension.

Generally, computing inconsistency scores on all possible pairs is expensive because the number of pairs grows quaratically w.r.t. $N$. Nevertheless, almost computational burden is attributed to operations of feature extraction network $f_\theta$ which composes convolutional layers. In pairwise manner, one patch is paired with other $N-1$ patches and passed through $f_\theta$ about $N-1$ times. This redundancy can be reduced. We first pre-extract low-dimensional features of all patches by evaluating $f_\theta (\mathcal{P}_i)$, $1 \leq i \leq N$, and proceed to compute $p_{ij}$ for all possible pairs using all computed features. 

For each patch $\mathcal{P}_k$ within the image, an inconsistency map $\mathcal{I}^{k} \in \mathbb{R}^ {N_H \times N_W}$ is built. If we consider all patches according to their spatial location on the image, $\mathcal{I}^k_{ij}$ is the inconsistency of $(i,j)$-th patch and $\mathcal{P}_k$, where $1 \leq i \leq N_H$ and $1 \leq j \leq N_W$.

\begin{figure}
	\includegraphics[width=\linewidth]{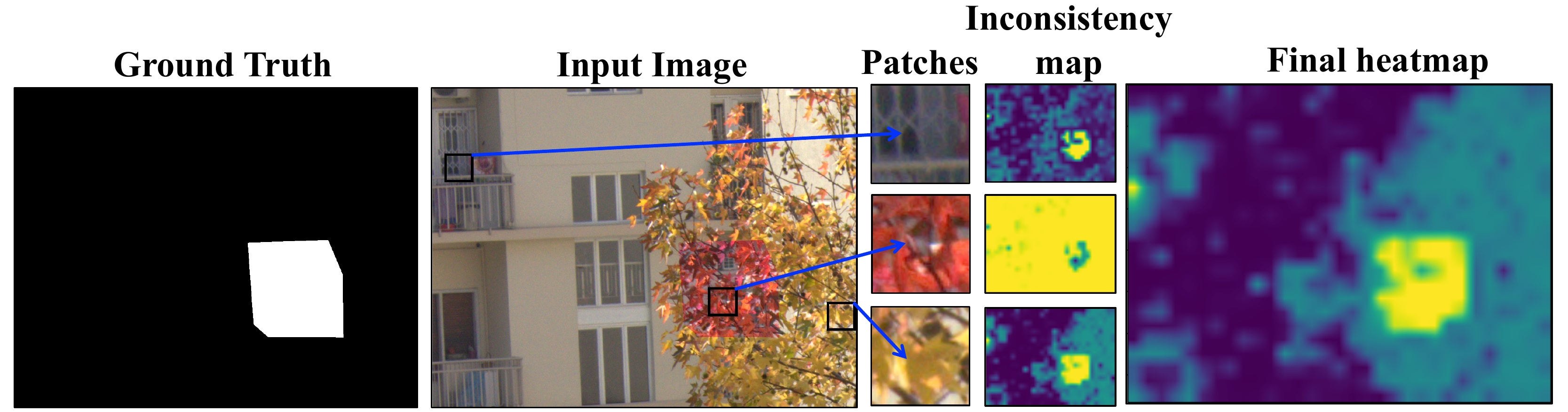}
	\caption{Fusing patch-level inconsistency maps into an  image-level map.}
	\label{fig:hm}
	\vspace{-0.5cm}
\end{figure}

It is typical to assume that the forged region is relatively small compared to the background, thus majority of $\mathcal{I}^{k}$ ($k$ refers to patches on the pristine region) exposes inconsistencies with the forged region, while remaining maps expose inconsistencies with the pristine region, as shown in Figure \ref{fig:hm}. In order to fuse inconsistency maps of majority patches belonging to the pristine region to obtain a unique map $\mathcal{\bar{I}} \in \mathbb{R}^ {N_H \times N_W}$, we follow the approach in \cite{Huh2018}, computing $\mathcal{\bar{I}}$ by mean shift algorithm\cite{Cheng1995}, which iteratively finds mean of majority (mode).
Eventually, $\mathcal{\bar{I}}$ is a subsampled heatmap which potentially highlights malicious region. The full-size heatmap can be obtained  by resizing $\mathcal{\bar{I}}$ with bilinear interpolation. If the forged region is larger than the background, we  obtain the inverted heatmap since the background is the smaller area. 
\subsubsection{Postprocessing} \label{sec:pp}
\begin{figure}[h]
	\centering
	\includegraphics[width=0.8\linewidth]{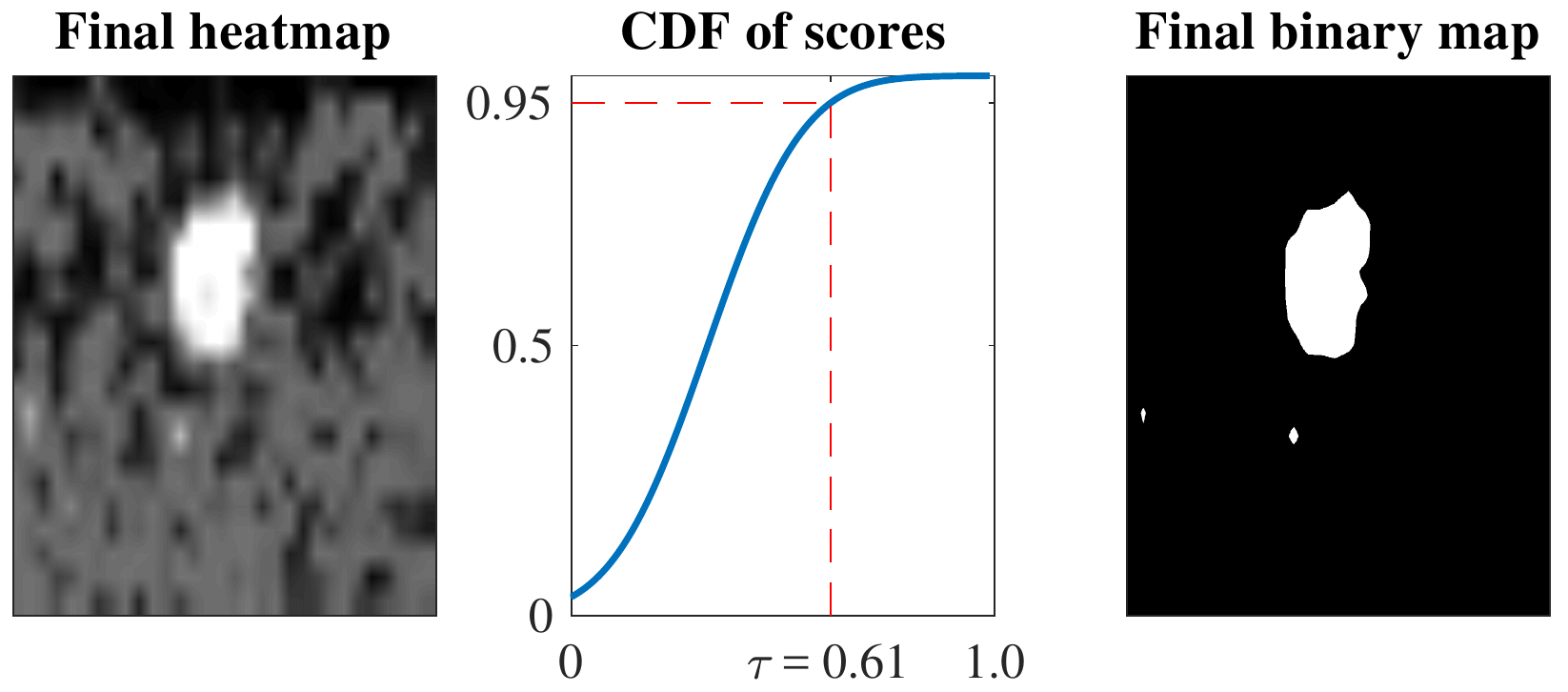}
	\caption{The heatmap before (left) and after (right) being postprocesed. The middle plot presents CDF of $\mathcal{\bar{I}}_{ij}$  and the threshold $\tau=0.61$.}
	\label{fig:pp}
	\vspace{-0.2cm}
\end{figure}

The standard logistic output $p_{ij}$ can be interpreted as the posterior probability that two patches $\mathcal{P}_i$ and $\mathcal{P}_j$ are inconsistent. After mean shifting, each element $\bar{p}_{ij} = \mathcal{\bar{I}}_{ij}$ tells us how probable $(i,j)$-th patch is forged because $\mathcal{\bar{I}}$ is the representative inconsistency map of pristine patches to all patches. While the threshold $0.5$ may be a reasonable choice for deciding if two patches are inconsistent, it is not straightfoward to apply this rule to pixel-level predictions. Moreover, as keeping False Alarm Rate (FAR) low is critial in forensic applications, a postprocessing step is important for pixel-level prediction. With this respect, postprocessing on each image is cast to finding a statistical threshold $\tau$ based on which a pixel is masked as forged or pristine. We apply a simple postprocessing based on the assumption that $\mathcal{\bar{I}}_{ij}$ (to avoid adding new notation, we mean $\mathcal{\bar{I}}$ after resized) follows a Gaussian distribution, $\mathcal{\bar{I}}_{ij} \sim N(\mu, \sigma^2)$. We fix $\tau$ such that $5\%$ of the right tail are decided as being forged. $\tau$ is lower bounded by $0.5$ to maintain acceptable FAR, namely $\tau = \max \left( 0.5, t  \right)$. $t$ is the solution of:
$	0.95 = \frac{1}{\sigma \sqrt{2\pi}}  \int_{-\infty}^{t} e^{\frac{-(x-\mu)^2}{2\sigma^2}} dx \text{.} \nonumber$
Compared to the threshold $0.5$, $\tau$ results in better or equal FAR. An example of postprocessing is shown in Figure \ref{fig:pp}.
\section{Experiments}\label{sec:exp}
Towards experimental evidences, we  evaluate our approach under different configurations and test sets.
\subsection{Test set}
To the best of our knowledge, there is no publicly available dataset on the problem of hue modification. Thus, to evaluate our method, we generated the test set from $120$ raw images of an external Canon 600D camera (never appeared in training phase) having CFA pattern $GBRG$. Raw images are decoded by \textsf{dcraw} version $9.27$. For each image, a top-left region is cropped out such that $H=768$ and $W=1024$. The forged area follows random convex shape fixed within a $256 \times 256$ bounding box, which is positioned at random location on the image. Next, we perform hue modification on pixels inside the polygons and generate multiple test sets:
\begin{itemize}
	\item $\mathcal{D}^\alpha_{png}$: Uncompressed images are demosaiced from raw images by \textsf{dcraw} and subject to local hue modification. For each modification angle $\alpha \in [30,330]$ step $30$, hue modification is carried out on $120$ uncompressed images.
	\item $\mathcal{D}^{QF}_{b-jpg}$: Hue modification by different angles ($10$ images for each modification angle $\alpha \in [30,330]$ step $30$) are carried out on $110$ images, and $10$ images are unmodified. Afterwards, all images are compressed using quality factors $QF \in [55,100]$, step $5$.
	\item $\mathcal{D}^{QF}_{a-jpg}$: $120$ images are first compressed using $QF \in [55,100]$, step $5$. Afterwards, hue modification by different angles (one angle for $10$ images) are carried out on $110$ JPEG images, while the remaining $10$ images are unmodified. All of them are compressed again using the default quality factor $75$. By the second JPEG compression, $\mathcal{D}^{QF}_{a-jpg}$ is more challenging since the training images are only subject to single JPEG compression.
\end{itemize}
\subsection{Setups}
The performance of our method is compared with the following state of the art methods: {\choi} \cite{Choi2013}, based on CFA-based artifacts and explicitly designed for the estimation of hue modification, and SpliceBuster \cite{Cozzolino2015}, based on statistical features of rich models \cite{Fridrich2012} and selected for comparison since those features potentially capture local disturbances caused by local hue modification. We do not compare with \cite{Hou2014,Hou2017} given their strong assumption about the availability of the reference PRNU which is unrealistic in practical scenarios.

This work particularly focuses on the localization of hue modification rather than its estimation.  Choi \textit{et al.} \cite{Choi2013} is an estimator which potentially returns the modification angle by searching over a feasible range. To convert Choi \textit{et al.} into a localization method, we use a sliding window $35$ similarly to our method, and search the angle over $[0,359]$, step $8$. If the angle found is $0$ or $352$, the patch is marked as pristine. Choi \textit{et al.} therefore outputs a binary map. The other method, SpliceBuster \cite{Cozzolino2015}, returns the negative log-likelihood that a pixel is pristine. It means, a large value indicates high probability that a pixel is forged. We linearly scale the returned map into $[0,1]$ and apply the same postprocessing described in Section \ref{sec:pp} to get the binary map.
In order to demonstrate the advantage of our postprocessing, we also report performance of the proposed method when a simple thresholding is applied to binarize the heatmap. We empirically found that the threshold $0.8$ yields most acceptable results.

We aggregate True Positive (TP), True Negative (TN), False Positive (FP), and False Negative (FN) over all images and report average True Positive Rate (TPR), True Negative Rate (TNR) and F1 score.
\subsection{Quatitative Evaluation}
\subsubsection{Detection on uncompressed images}
\begin{figure}[h]
	\includegraphics[width=\linewidth]{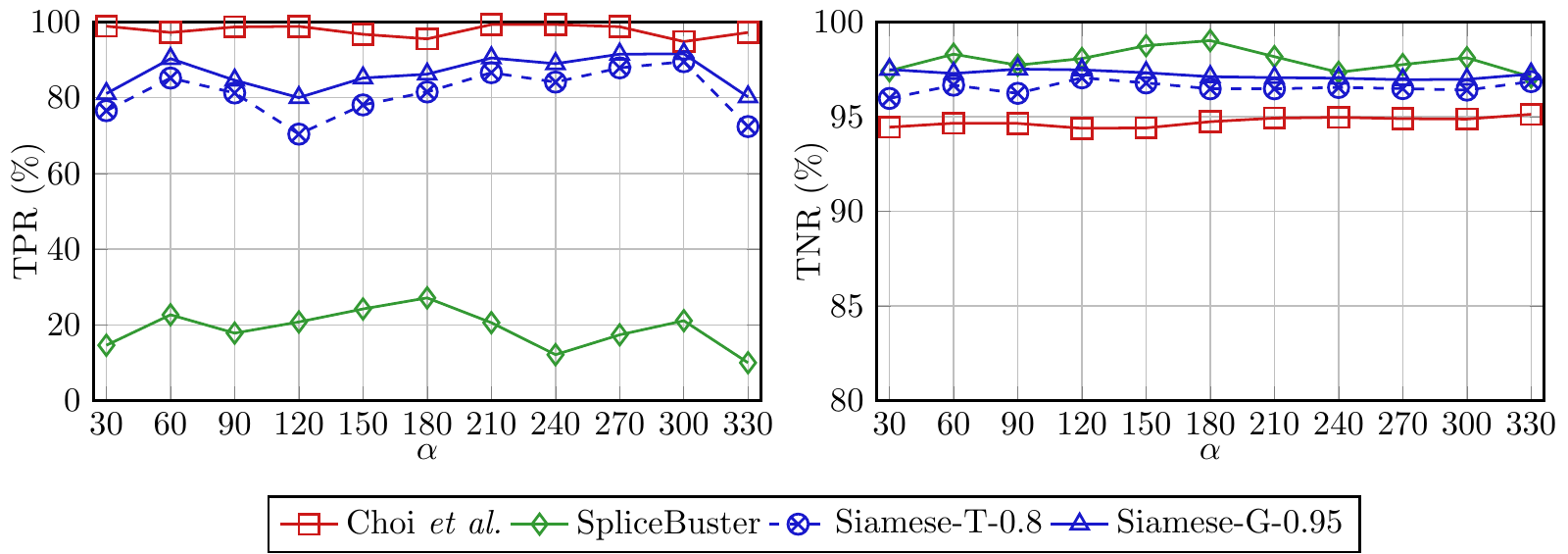}
	\caption{TPR and TNR of all methods on $\mathcal{D}^{\alpha}_{png}$.}
	\label{fig:png_tpr_tnr}
\end{figure}

In this section, we evaluate the first model trained on uncompressed images, and compare with {\choi} and SpliceBuster on $\mathcal{D}^\alpha_{png}$. Figure \ref{fig:png_tpr_tnr} presents TPR and TNR obtained by all methods. 
{\choi} is guaranteed to detect hue modification on uncompressed images since this type of manipulation distorts demosaicing artifacts. It achieves high TPR which implies that almost forged pixels are correctly detected. This comes at a cost of slightly worse TNR. SpliceBuster, on the other hand, detects correctly only about $20\%$ of forged pixels, and as a consequence, yields very high TNR. We assume that the  features used in \cite{Fridrich2012} are ineffective for hue modification detection.

Our Siamese network with heatmaps thresholded simply by $0.8$ is denoted by {\siamesethr}. The other alternative is denoted by {\siamesegau}, where heatmaps are postprocessed by threshold $\tau$ as designed in Section \ref{sec:pp}. We can clearly see that {\siamesegau} outperforms {\siamesethr} in all cases. In fact, a fixed threshold over all heatmaps cannot deal with high variability of predictive scores on each heatmap, and thus an adaptive threshold is more effective.
Interestingly, the TPR reveals the fact that the middle range of modification angles are easier to detect by our methodologies. This is explainable since the strength of hue modification is periodic with the period of $360^\circ$. Very small or very large positive angles correspond to little modifications.

We summarize the overall performance for some selective modification angles in Table \ref{tab:png_f1}. In terms of F1 score, {\siamesegau} outperforms all other methods.
\begin{table}
	\centering
	\caption{F1 scores of all methods on $\mathcal{D}^{\alpha}_{png}$.}
	\begin{tabular}{ C{1.85cm} | C{0.5cm} C{0.5cm} C{0.5cm} C{0.5cm} C{0.5cm} C{0.5cm} }
		Angle $\alpha \rightarrow$ Method $\downarrow$ & 30 & 90 & 150 & 210 & 270 & 330 \\
	\hline
	{\choi} & 66.16 & 67.15 & 65.05 & 68.34 & 67.92 & 68.21 \\
	SpliceBuster & 18.12 & 29.57 & 32.99 & 26.80 & 21.97 & 12.29 \\
	{\siamesethr} & 66.44 & 65.28 & 66.22 & 69.28 & 70.00 & 63.23 \\
	{\siamesegau} & \textbf{71.79} & \textbf{73.82} & \textbf{73.04} & \textbf{74.41} & \textbf{74.29} & \textbf{69.86} \\ 
	\end{tabular}
	\label{tab:png_f1}
	\vspace{-0.3cm}
\end{table}
\subsubsection{Detection in the presence of JPEG compression} We target more practical scenarios where hue modification is done with the presence of JPEG compression. It has been acknowledged that JPEG compression has strong impact on demosaicing artifacts \cite{Gallagher2008,Cao2009,Kirchner2010}. {\choi} is also very sensitive to JPEG compression since the count of interpolated and recorded pixels is less accurate \cite{Choi2013}.

We assess the second model trained on JPEG images under two testing circumstances: i) hue modification is performed on uncompressed images followed by JPEG compression, and ii) hue modification is performed on JPEG compressed images, and those are subsequently compressed again using quality factor $75$. Note that during training, we do not perform second JPEG compression.

\begin{figure}[h]
	\includegraphics[width=\linewidth]{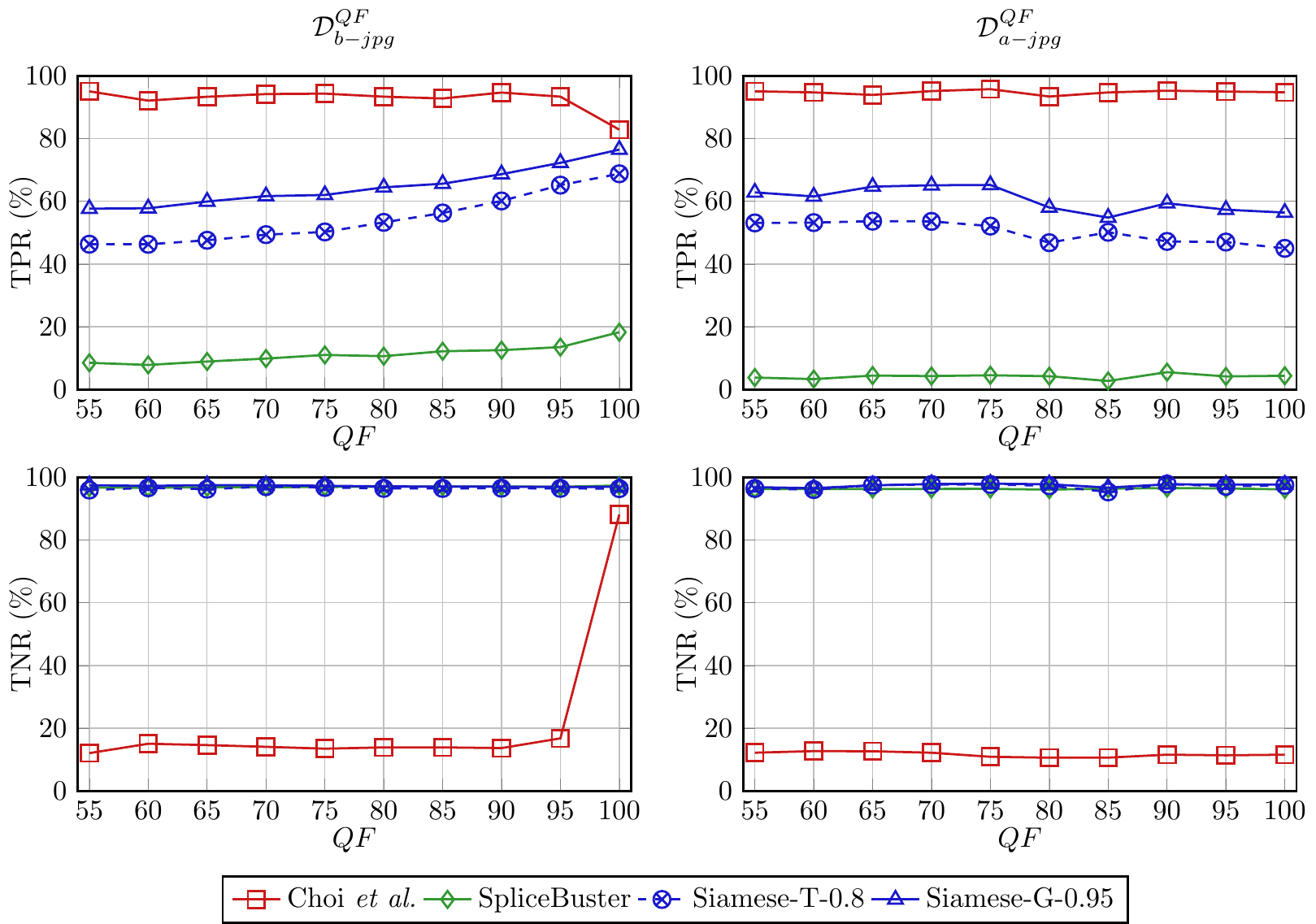}
	\caption{TPR and TNR of all methods on $\mathcal{D}^{QF}_{b-jpg}$ and $\mathcal{D}^{QF}_{a-jpg}$. $QF$ is the first compression quality factor.}
	\label{fig:jpg_tpr_tnr}
\end{figure}

The TPR and TNR of all methods, where hue modification is performed before JPEG compression, i.e. dataset $\mathcal{D}^{QF}_{b-jpg}$, is shown in the first column of Figure \ref{fig:jpg_tpr_tnr}. {\choi} fails to spot forged area unless the image is compressed with highest $QF$. At $QF=100$, {\choi} achieves TPR $82.88\%$ and TNR $88.15\%$. SpliceBuster, on the other hand, can only detect about $10\%$ of forged pixels. Our proposed methods perform far better than the other two competitors on $\mathcal{D}^{QF}_{b-jpg}$, by keeping TPR at acceptable level and retaining always high TNR, i.e.,  {\siamesegau} achieves an average $64.67\%$ of TPR ($\approx 65\%$ of forged pixels are correctly detected) and $97.72\%$ of TNR, while {\siamesethr} attains $54.35\%$ of TPR and $97.39\%$ of TNR. 
In the right column of Figure \ref{fig:jpg_tpr_tnr}, i.e. dataset $\mathcal{D}^{QF}_{a-jpg}$, the overall TPR and TNR of our methods are slightly degraded compared to the performance on $\mathcal{D}^{QF}_{b-jpg}$. This degradation can be attributed to the second JPEG compression. In fact, we can generally observe the correlation of performance degradation and compression rate: the higher the first $QF$, the lower the performance. While {\choi} behaves positively on $\mathcal{D}^{100}_{b-jpg}$, it loses that capability on $\mathcal{D}^{100}_{a-jpg}$.
\begin{table}[t!]
	\centering
	\caption{F1 scores of all methods on $\mathcal{D}^{QF}_{b-jpg}$.}
	\begin{tabular}{ C{1.85cm} | C{0.5cm} C{0.5cm} C{0.5cm} C{0.5cm} C{0.5cm} C{0.5cm} }
		$QF$ $\rightarrow\;\;\;$ Method $\downarrow$ & 75 & 80 & 85 & 90 & 95 & 100 \\
	\hline
	{\choi} & 9.80 & 9.74 & 9.69 & 9.85 & 10.05 & 39.47 \\
	SpliceBuster & 12.80 & 12.47 & 14.26 & 14.66 & 15.57 & 21.44 \\
	{\siamesethr} & 51.96 & 53.43 & 54.77 & 54.56 & 56.81 & 61.60 \\
	{\siamesegau} & \textbf{61.41} & \textbf{60.83} & \textbf{63.03} & \textbf{64.65} & \textbf{66.73} & \textbf{69.11} \\ 
	\end{tabular}
	\label{tab:b_jpg_f1}
	\vspace{-0.4cm}
\end{table}

The overall F1 scores for several selective $QF$ are shown in Table \ref{tab:b_jpg_f1} and \ref{tab:a_jpg_f1}. Our two methodologies, in particular {\siamesegau}, outperform the other two methods to a large margin. We might notice that {\choi} achieves F1 score $39.47\%$ on $\mathcal{D}^{100}_{b-jpg}$ while TPR and TNR in the same configuration are over $80\%$, see left column in Figure \ref{fig:jpg_tpr_tnr}. This phenomenon is due to the high FP which penalizes precision, and as a consequence, F1 score. However, since TN dominates FP (due to the large pristine area compared to the forged area), TNR is not effectively penalized.
\begin{table}[t!]
	\centering
	\caption{F1 scores of all methods on $\mathcal{D}^{QF}_{a-jpg}$.}
	\begin{tabular}{ C{1.85cm} | C{0.5cm} C{0.5cm} C{0.5cm} C{0.5cm} C{0.5cm} C{0.5cm} }
		$QF$ $\rightarrow\;\;\;$ Method $\downarrow$ & 75 & 80 & 85 & 90 & 95 & 100 \\
	\hline
	{\choi} & 9.68 & 9.62 & 9.55 & 9.70 & 9.65 & 9.65 \\
	SpliceBuster & 5.07 & 4.63 & 3.08 & 6.35 & 4.74 & 4.82 \\
	{\siamesethr} & 52.85 & 46.76 & 41.29 & 49.93 & 46.13 & 46.09 \\
	{\siamesegau} & \textbf{63.51} & \textbf{57.52} & \textbf{49.83} & \textbf{58.36} & \textbf{56.32} & \textbf{55.75} \\ 
	\end{tabular}
	\label{tab:a_jpg_f1}
	\vspace{-0.4cm}
\end{table}
\subsection{Qualitative Inspection}
\begin{figure}[t!]
	\centering
	\includegraphics[width=\linewidth]{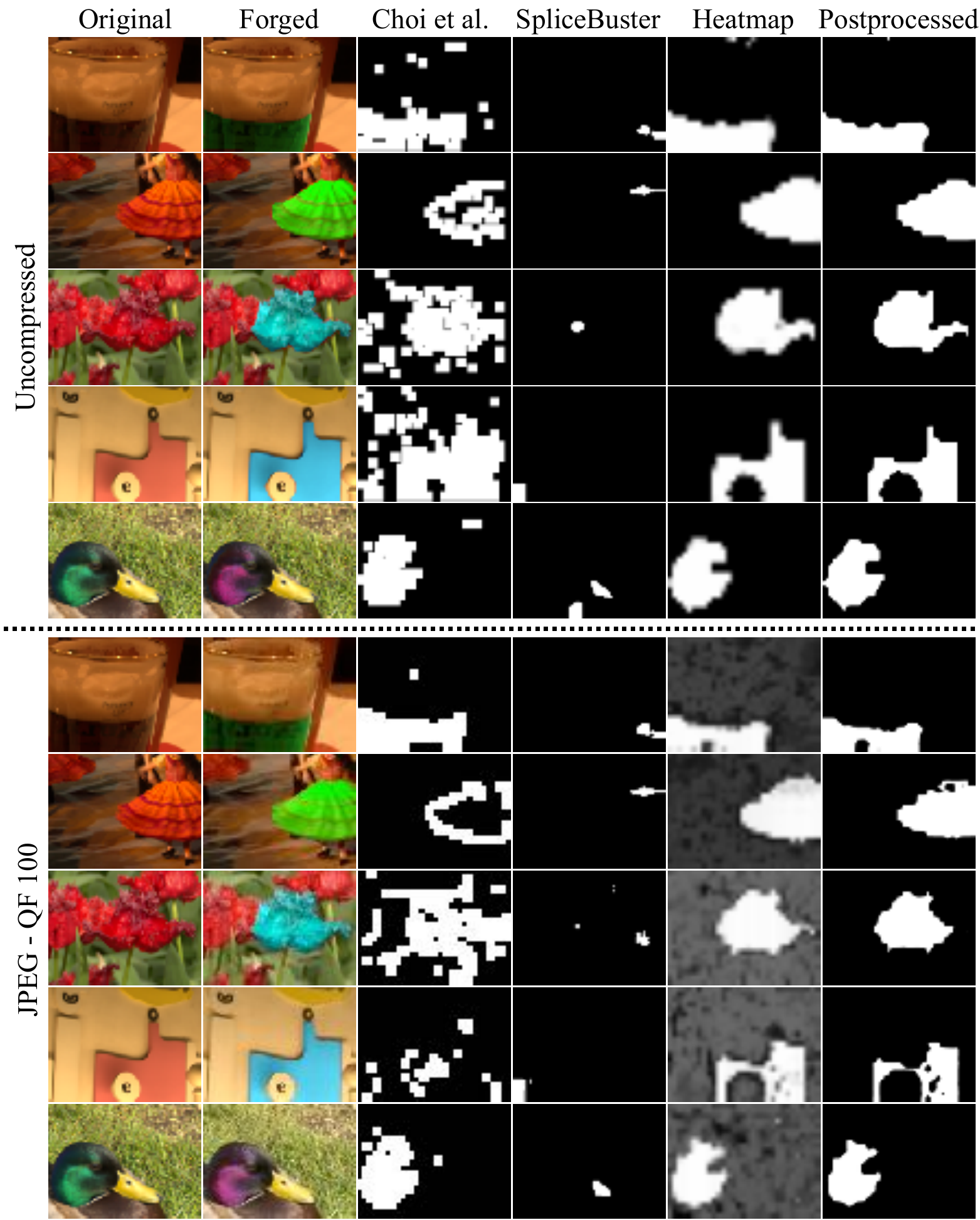}
	\caption{Detection results on realistic examples.}
	\label{fig:realistic_examples}
	\vspace{-0.5cm}
\end{figure}

In Figure \ref{fig:realistic_examples}, we provide detection results on realistic examples manually created using GIMP. Hue modification is carried out on uncompressed images (the first $5$ lines) and the modified images are JPEG compressed using highest quality $QF=100$ (the last $5$ lines). {\siamesegau} (the last column) clearly results in better detection maps compared with {\choi} and SpliceBuster.
\section{Conclusion}\label{sec:con}
We have proposed a data-driven countermeasure for hue modification on color images based on patch matching. This task is done by means of a Siamese architecture which receives the two inputs and outputs the likelihood that the two inputs are inconsistent. A unique localization map is generated from inconsistency scores of multiple patches. Our models perform well on uncompressed and JPEG compressed images even though JPEG compression distorts CFA and demosaicing artifacts. Our future investigations will focus on the estimation of hue modification angles, based on which the original image can be recovered.
\bibliographystyle{IEEEtran}
\bibliography{references}

\begin{thebibliography}{10}
\providecommand{\url}[1]{#1}
\csname url@samestyle\endcsname
\providecommand{\newblock}{\relax}
\providecommand{\bibinfo}[2]{#2}
\providecommand{\BIBentrySTDinterwordspacing}{\spaceskip=0pt\relax}
\providecommand{\BIBentryALTinterwordstretchfactor}{4}
\providecommand{\BIBentryALTinterwordspacing}{\spaceskip=\fontdimen2\font plus
\BIBentryALTinterwordstretchfactor\fontdimen3\font minus
  \fontdimen4\font\relax}
\providecommand{\BIBforeignlanguage}[2]{{%
\expandafter\ifx\csname l@#1\endcsname\relax
\typeout{** WARNING: IEEEtran.bst: No hyphenation pattern has been}%
\typeout{** loaded for the language `#1'. Using the pattern for}%
\typeout{** the default language instead.}%
\else
\language=\csname l@#1\endcsname
\fi
#2}}
\providecommand{\BIBdecl}{\relax}
\BIBdecl

\bibitem{Ferrara2012}
P.~{Ferrara}, T.~{Bianchi}, A.~{De Rosa}, and A.~{Piva}, ``Image forgery
  localization via fine-grained analysis of {CFA} artifacts,'' \emph{IEEE
  Trans. on Information Forensics and Security}, vol.~7, no.~5, pp. 1566--1577,
  2012.

\bibitem{Fridrich2012}
J.~{Fridrich} and J.~{Kodovsky}, ``Rich models for steganalysis of digital
  images,'' \emph{IEEE Trans. on Information Forensics and Security}, vol.~7,
  no.~3, pp. 868--882, 2012.

\bibitem{Cozzolino2015}
D.~{Cozzolino}, G.~{Poggi}, and L.~{Verdoliva}, ``Splicebuster: A new blind
  image splicing detector,'' in \emph{Proc. of WIFS}, 2015, pp. 1--6.

\bibitem{Li2017ForgeryLocalization}
H.~{Li}, W.~{Luo}, X.~{Qiu}, and J.~{Huang}, ``Image forgery localization via
  integrating tampering possibility maps,'' \emph{IEEE Trans. on Information
  Forensics and Security}, vol.~12, no.~5, pp. 1240--1252, 2017.

\bibitem{Chen2008}
M.~Chen, J.~Fridrich, M.~Goljan, and J.~Luk\'{a}\v{s}, ``Determining image
  origin and integrity using sensor noise,'' \emph{IEEE Trans. on Information
  Forensics and Security}, vol.~3, no.~1, pp. 74--90, 2008.

\bibitem{Choi2013}
C.-H. Choi, H.-Y. Lee, and H.-K. Lee, ``Estimation of color modification in
  digital images by {CFA} pattern change,'' \emph{Forensic science
  international}, vol. 226, pp. 94--105, 01 2013.

\bibitem{Hou2014}
J.~{Hou}, H.~{Jang}, and H.~{Lee}, ``Hue modification estimation using sensor
  pattern noise,'' in \emph{ICIP}, 2014, pp. 5287--5291.

\bibitem{Hou2017}
J.~Hou and H.~Lee, ``Detection of hue modification using photo response
  nonuniformity,'' \emph{IEEE Trans. on Circuits and Systems for Video
  Technology}, vol.~27, no.~8, pp. 1826--1832, 2017.

\bibitem{Bromley1993}
J.~Bromley, I.~Guyon, Y.~LeCun, E.~S\"{a}ckinger, and R.~Shah, ``Signature
  verification using a `'siamese`' time delay neural network,'' in \emph{Proc.
  of NIPS}, 1993, pp. 737--744.

\bibitem{Cozzolino2018NoisePrint}
\BIBentryALTinterwordspacing
D.~Cozzolino and L.~Verdoliva, ``Noiseprint: a {CNN}-based camera model
  fingerprint,'' \emph{CoRR}, vol. abs/1808.08396, 2018. [Online]. Available:
  \url{http://arxiv.org/abs/1808.08396}
\BIBentrySTDinterwordspacing

\bibitem{Mayer2018}
O.~Mayer and M.~C. Stamm, ``Learned forensic source similarity for unknown
  camera models,'' in \emph{Proc. of ICASSP}, 2018, pp. 2012--2016.

\bibitem{Huh2018}
M.~Huh, A.~Liu, A.~Owens, and A.~A. Efros, ``Fighting fake news: Image splice
  detection via learned self-consistency,'' in \emph{Proc. of The ECCV}, 2018.

\bibitem{He2016}
K.~He, X.~Zhang, S.~Ren, and J.~Sun, ``Deep residual learning for image
  recognition,'' in \emph{Proc. of CVPR}, 2016, pp. 770--778.

\bibitem{Dang-Nguyen2015}
D.-T. Dang-Nguyen, C.~Pasquini, V.~Conotter, and G.~Boato, ``{RAISE}: A raw
  images dataset for digital image forensics,'' in \emph{Proc. of ACM MMSys},
  2015, pp. 219--224.

\bibitem{Gloe2010}
T.~Gloe and R.~B\"{o}hme, ``The `{D}resden {I}mage {D}atabase' for benchmarking
  digital image forensics,'' in \emph{Proc. of ACM SAC}, vol.~2, 2010, pp.
  1585--1591.

\bibitem{Deng2009}
J.~{Deng}, W.~{Dong}, R.~{Socher}, L.~{Li}, and and, ``{ImageNet}: A
  large-scale hierarchical image database,'' in \emph{2009 IEEE Conference on
  Computer Vision and Pattern Recognition}, 2009, pp. 248--255.

\bibitem{Cheng1995}
Y.~Cheng, ``Mean shift, mode seeking, and clustering,'' \emph{IEEE Trans. on
  Pattern Analysis and Machine Intelligence}, vol.~17, no.~8, pp. 790--799,
  1995.

\bibitem{Gallagher2008}
A.~C. {Gallagher} and T.-H. Chen, ``Image authentication by detecting traces of
  demosaicing,'' in \emph{Proc. of IEEE Workitorial on Vision of the Unseen (in
  conjunction with CVPR)}, 2008, pp. 1--8.

\bibitem{Cao2009}
H.~Cao and A.~C. Kot, ``Accurate detection of demosaicing regularity for
  digital image forensics,'' \emph{IEEE Transactions on Information Forensics
  and Security}, vol.~4, no.~4, pp. 899--910, 2009.

\bibitem{Kirchner2010}
M.~{Kirchner}, ``Efficient estimation of {CFA} pattern configuration in digital
  camera images,'' in \emph{Proc. of SPIE, Media Forensics and Security II},
  vol. 7541, 2010.

\end{thebibliography}

\end{document}